\documentclass[twoside]{article}

\usepackage{aistats2020}
\usepackage[utf8]{inputenc} 
\usepackage[T1]{fontenc}    
\usepackage{hyperref}       
\usepackage{url}            
\usepackage{booktabs}       
\usepackage{nicefrac}       
\usepackage{microtype}      
\usepackage{natbib}
\usepackage{float}
\usepackage{amsmath, amsfonts, amssymb}
\usepackage{color}
\usepackage{graphicx}
\usepackage{makecell}
\usepackage{array, multirow}
\usepackage{adjustbox}

\begin{document}

%

%

\twocolumn[

\aistatstitle{On the expected behaviour of noise regularised \\deep neural networks as Gaussian processes}

\aistatsauthor{ Arnu Pretorius \And Herman Kamper \And  Steve Kroon }

\aistatsaddress{ Computer Science Division \\
Stellenbosch University \And  Department of E\&E Engineering \\
Stellenbosch University \And Computer Science Division \\
Stellenbosch University } 
]

\begin{abstract}
    Recent work has established the equivalence between deep neural networks and Gaussian processes (GPs), resulting in so-called \textit{neural network Gaussian processes} (NNGPs). 
    The behaviour of these models depends on the initialisation of the corresponding network. 
    In this work, we consider the impact of noise regularisation (e.g. dropout) on NNGPs, and relate their behaviour to signal propagation theory in noise regularised deep neural networks. 
    For ReLU activations, we find that the best performing NNGPs have kernel parameters that correspond to a recently proposed initialisation scheme for noise regularised ReLU networks. 
    In addition, we show how the noise influences the covariance matrix of the NNGP, producing a stronger prior towards simple functions away from the training points. 
    We verify our theoretical findings with experiments on MNIST and CIFAR-10 as well as on synthetic data.
\end{abstract}
\section{Introduction}

\begin{figure*}
    \centering
    \includegraphics[width=0.9\linewidth]{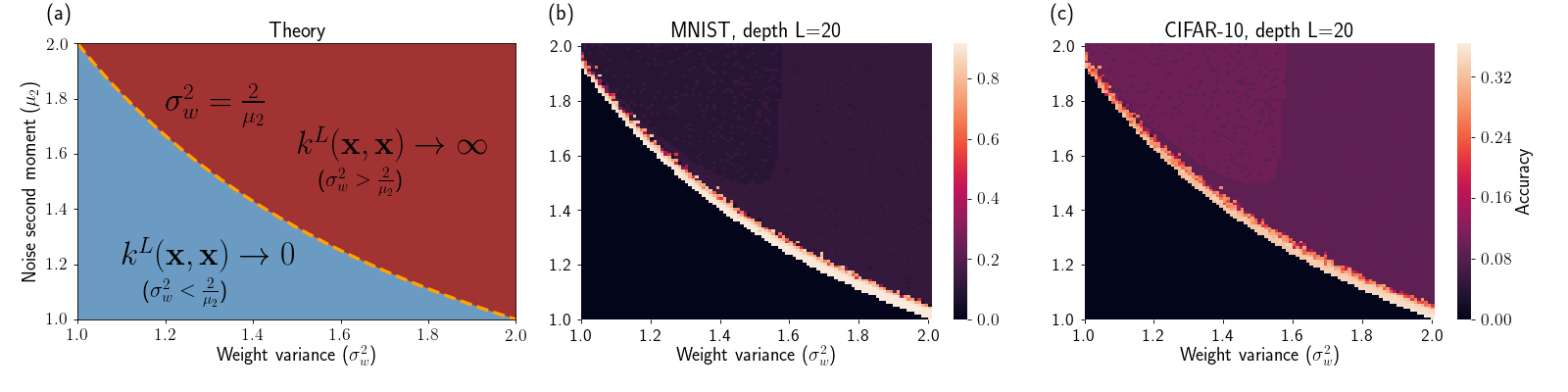}
    \vspace{-0.5cm}
	\caption{\textit{Dependence of noisy NNGPs on critical parameters for performance.} \textbf{(a)} Critical boundary for kernel parameters $\{\sigma^2_w, \sigma^2_b\} = \{2/\mu_2, 0\}$ as a function of noise.  \textbf{(b)} MNIST test accuracy for a $20$-layer noisy NNGP, for kernel parameters $\sigma^2_b = 0$, $\sigma^2_w, \mu_2 \in [1.0, 2.0]$ (both sampled in interval sizes of $0.01$). Training and test set sizes are $N=1000$. \textbf{(c)} CIFAR-10 test accuracy, details same as (b).} 
	\label{fig: nggp performance}
\end{figure*}

Modern deep neural networks (NNs) are powerful tools for modeling highly complex functions. 
However, deep NNs lack natural ways of incorporating uncertainty estimation, and (approximate) Bayesian inference for NNs remains a challenge. 
In contrast, non-parameteric approaches such as Gaussian Processes (GPs) provide exact Bayesian inference and well-calibrated uncertainty estimates, but typically consider substantially simpler models than deep NNs. 
Therefore, a large body of work has recently emerged attempting to combine parametric deep learning models and GPs so as to derive benefits from both. 
These approaches include deep GPs \citep{damianou2013deep,duvenaud2014avoiding, hensman2014nested,dai2015variational, bui2016deep, salimbeni2017doubly}, deep kernel learning \citep{wilson2016deep,wilson2016stochastic,al2016learning} and viewing deep learning with dropout as an approximate deep GP \citep{gal2016dropout}.

For shallow infinite width neural networks, an exact equivalence to GPs has been known for some time \citep{neal1994priors, williams1997computing, le2007continuous}. However, this equivalence has only recently been extended to deeper architectures \citep{hazan2015steps, lee2017deep, matthews2018gaussian, novak2018bayesian}. Referred to as \textit{neural network Gaussian processes} (NNGPs) in \citet{lee2017deep}, the resulting models are GPs with an exact correspondence to infinitely wide deep neural networks. Importantly, the NNGP depends on the hyperparameters of the network and its initialisation, which determines the network's signal propagations dynamics. 


In deep neural networks, signal propagation has been shown to exhibit distinct phases depending on the initialisation of the network \citep{poole2016exponential}. 
These phases include ordered and chaotic regimes associated with vanishing and exploding gradients respectively, which can result in poor network performance \citep{schoenholz2016deep}. 
By initialising at the critical boundary between these two regimes, known as the ``edge of chaos'', the flow of information through the network improves, often resulting in faster and deeper training for a variety of architectures \citep{pennington2017resurrecting, yang2017mean, chen2018dynamical,xiao2018dynamical}. 

\citet{lee2017deep} showed that NNGPs tend to inherit the above behaviour from their corresponding randomly initialised networks. 
In particular, there exists an interaction between poor signal propagation and a poorly constructed kernel. 
As a result, the performance of NNGPs tend to suffer if their kernels are constructed using kernel parameters that correspond to network initialisations far from the critical boundary. 
Furthermore, even at the critical boundary, inputs to a neural network can still become asymptotically correlated at large depth \citep{schoenholz2016deep}. 
The rate of convergence in this correlation limits the depth to which networks can be trained, because after this convergence the network is unable to distinguish between different training observations at the output layer. 
This dependence on depth (in the constructed kernel) for performance, also manifests in NNGPs \citep{lee2017deep}.


Various design decisions are required to instantiate a modern NN. 
Important decisions for trainability and test performance often include both initialisation \textit{and} regularisation.
If initialised poorly, a network might become untrainable due to poor signal propagation, whereas a lack of regularisation could hurt the test performance of the network if it starts to overfit. 
Commonly used approaches to alleviating these issues include principled initialisation schemes \citep{glorot2010understanding, he2015delving} and improved regularisation strategies. 
Among the most successful regularisation strategies is dropout \citep{srivastava2014dropout}, a form of noise regularisation where scaled Bernoulli noise is applied multiplicatively to the units of a network to prevent co-adaptation. 
However, as shown by \citet{pretorius2018critical}, these components do not act in isolation and therefore the initialisation of the network should depend on the amount of noise regularisation being applied.


In this paper, we investigate the following research question: do the signal propagation dynamics that influence noise regularised NNs also govern the behaviour of corresponding ``noisy NNGPs''? 
In the presence of multiplicative noise regularisation, \citet{pretorius2018critical} derived the critical initialisation for stable signal propagation in feedforward ReLU networks: More specifically, the authors showed that stable propagation is achieved by setting all unit biases to zero and sampling the weights from zero mean Gaussians with variance $\sigma^2_w/D_{\textrm{in}}$ set equal to $\sigma^2_w = 2/\mu_2$. Here, $D_{\textrm{in}}$ is the number of incoming units to the layer and $\mu_2$ is the second moment of the noise. 
For example, when using dropout, $\mu_2 = 1/p$ (where $p$ is the probability of keeping the unit active) and therefore the initial weights are sampled from $\mathcal{N}(0, 2p/D_{\textrm{in}})$.
Furthermore, it was shown that the rate of convergence to a fixed point correlation between inputs increases as a function of the amount of regularisation being applied. 
Consequently, increased noise further limits the depth of trainability in neural networks. In this paper, we investigate whether these findings for noise regularised networks carry across to their noisy NNGP counterparts. 

We consider noise regularised fully-connected feedforward NNs and study the behaviour of noisy NNGPs. Our analysis is done in expectation over the noise, under a general noise model (of which dropout is a special case). We give the kernel corresponding to noisy ReLU NNGPs and highlight the different noise-induced degeneracies in the kernel as the depth becomes large. Specifically, we show that the above noise dependent initialisations promoting stable signal propagation in noisy ReLU NNs correspond exactly to the kernel parameters exhibiting good performance in noisy NNGPs, as shown in Figure \ref{fig: nggp performance}. However, even at criticality, we show that as the noise tends to infinity the covariance of the NNGP becomes diagonal.
As a result, noise regularisation translates into a stronger prior for simple functions away from the training points.
Finally, we verify our findings with experiments on real-world and synthetic datasets. 


\section{Noise regularised deep neural networks as Gaussian processes}


\begin{figure*}
    \centering
    \includegraphics[width=0.7\linewidth]{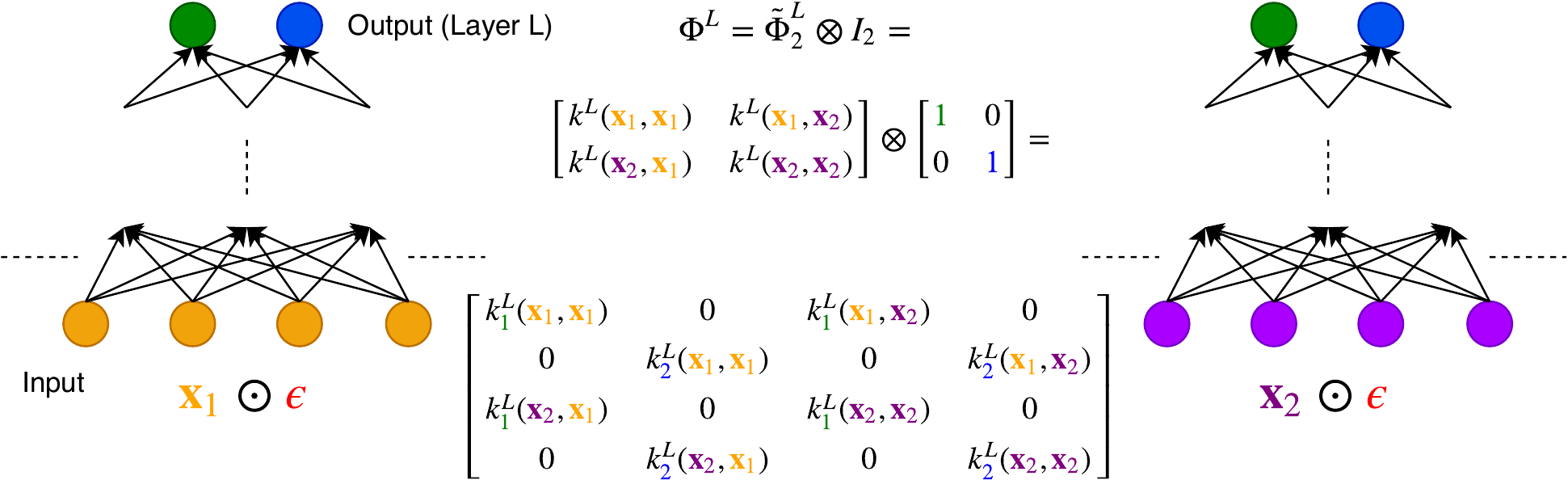}
	\caption{\textit{Noisy NNGP covariance}: Example of the covariance for a noisy NNGP with only two inputs $\mathbf{x}_1$ (orange), $\mathbf{x}_2$ (purple) and two output units (green and blue) at layer $L$ and sampled noise $\epsilon$.} 
	\label{fig: nggp covariance}
\end{figure*}

We consider a noise regularised fully-connected deep feedforward neural network. Given an input $\mathbf{x}^0 \in \mathbb{R}^{D_0}$, we inject noise into the model
\begin{align}
	\tilde{\mathbf{h}}^l = W^l(\mathbf{x}^{l-1}\odot \epsilon^{l-1}) + \mathbf{b}^l, \textcolor{white}{spa} \text{ for } l = 1, ..., L
	\label{eq: noisy deep model}
\end{align}
using the operator~$\odot$ to denote either addition or multiplication, where $\epsilon^l$ is an input noise vector, sampled from a pre-specified noise distribution.
The noise is assumed to have $\mathbb{E}[\epsilon^l] = 0$ in the additive case, and $\mathbb{E}[\epsilon^l] = 1$ for multiplicative noise distributions such that in both cases, $\mathbb{E}_\epsilon[\tilde{\mathbf{h}}^l] = W^l\mathbf{x}^{l-1} + \mathbf{b}^l$. 
The weights $W^l \in \mathbb{R}^{D_l \times D_{l-1}}$ and biases $\mathbf{b}^l \in \mathbb{R}^{D_l}$ are sampled i.i.d.\ from zero mean Gaussian distributions with variances $\sigma^2_w/D_{l-1}$ and $\sigma^2_b$, respectively, where $D_l$ denotes the dimensionality of the $l^{th}$ hidden layer in the network.
Each hidden layer's activations $\mathbf{x}^l = \phi(\tilde{\mathbf{h}}^l)$ are computed element-wise using an activation function $\phi(\cdot)$. 
Lastly, we denote the second moment of the noise as $\mu_2 = \mathbb{E}[\epsilon^2]$ and define $\tilde{\mathbf{h}}^L = f(\mathbf{x})$ as the entire function mapping from input to output, with $\mathbf{x} = \mathbf{x}^0$. 


By choosing parameter sampling distributions at initialisation we are implicitly specifying a prior over networks in parameter space. 
We now transition from parameter space to function space by instead specifying a prior directly over function values. 
Assume a training set of input-output pairs $\mathcal{D} = \{(\mathbf{x}_i, y_i), i = 1, ..., N\}$. If we can show that $\mathbf{f} = [f(\mathbf{x}_1), ..., f(\mathbf{x}_N)]^T$ is Gaussian distributed at initialisation, then the distribution over the output of the network at these points is completely determined by the second-order statistics $\mathbb{E}[\mathbf{f}] = \mathbf{m}$ and $\text{Cov}[\mathbf{f}]=K$, defining the following GP 
\begin{align}
    \label{eq: gp def}
    p(\mathbf{f}) = \mathcal{N}(\mathbf{m}, K) \equiv \mathcal{GP}(\mathbf{m}, K)  .
\end{align}
We begin by assuming the following additive error model with regression outcomes $y_i = f(\mathbf{x}_i) + \varepsilon_i$, where $\varepsilon_i \sim \mathcal{N}(0, \sigma^2_\varepsilon)$.\footnote{Note that here we consider scalar outputs, i.e. $f: \mathbb{R}^{D_0} \rightarrow \mathbb{R}$, hence $\tilde{h}^L = f(\mathbf{x}) \in \mathbb{R}$. Also, the additive error noise $\varepsilon$ should not be confused with the injected noise $\epsilon^l$ in \eqref{eq: noisy deep model}.} Since $y_i | \mathbf{x}_i \sim \mathcal{N}(f(\mathbf{x}_i), \sigma^2_\varepsilon)$, the joint distribution over all outcomes is
\begin{align}
    \label{eq: conditional for y}
    p(\mathbf{y}|\mathbf{f}) = \mathcal{N}(\mathbf{f}, \sigma^2_\varepsilon\mathbf{I}_N),
\end{align}
where $\mathbf{y} = [y_1, ..., y_N]^T$. In GP regression we are interested in finding the marginal distribution
\begin{align}
    \label{eq: marginal of y}
    p(\mathbf{y}) = \int p(\mathbf{y}|\mathbf{f})p(\mathbf{f})d\mathbf{f}.
\end{align}
We proceed as in \citep{lee2017deep} to argue that $\mathbf{f}$ is in fact a zero mean Gaussian (we refer the reader to \cite{matthews2018gaussian} for a more formal approach) and derive the elements of the covariance matrix $K$ in \eqref{eq: gp def} for noise regularised deep neural networks. 
Subsequently, we obtain an expression for \eqref{eq: marginal of y} by combining \eqref{eq: gp def} and \eqref{eq: conditional for y} and using standard results for the marginal of a Gaussian. 

To show that the expected distribution of $\mathbf{f}$ over the injected noise is Gaussian, we first note that conditioned on the inputs, the ``output'' units at layer $l$, stemming from the post-activations $\mathbf{x}^{l-1}$ in the previous layer are given by $\tilde{h}_d^l = \mathbf{w}^l_d \cdot (\mathbf{x}^{l-1}\odot \epsilon^{l-1}) + b^l_d$, for $d = 1, ..., D_l$. 
As previously mentioned, we sample the weights and biases i.i.d.\ from a zero mean Gaussian and define the noise to be i.i.d.\ such that $\tilde{h}^l_d$ is unbiased in expectation of the injected noise.
Therefore, in a wide network, $\tilde{h}^l_d$ is a sum of a large collection of i.i.d.\ random variables. 
As $D_{l-1} \rightarrow \infty$, the central limit theorem ensures that the distribution of $\tilde{h}^l_d$ will tend to a Gaussian with mean $\mathbb{E}[\tilde{h}^l_d(\mathbf{x}_i)] = 0$ and covariance $\mathbb{E} \left[ \tilde{h}^l_d(\mathbf{x}_i) \tilde{h}^l_d(\mathbf{x}_j) \right]$.
As a result, the function values $h^l_d(\mathbf{x}_1), ..., h^l_d(\mathbf{x}_N), \forall d$ can be treated as samples from a GP given by $\tilde{\mathbf{h}}^l \sim \mathcal{GP}(\mathbf{0}, \Phi^l)$. 
Here, $\Phi^l$ is an $ND_l \times ND_l$ covariance matrix given by
\begin{align*}
    \Phi^l & = \begin{bmatrix}
        k^l(\mathbf{x}_1, \mathbf{x}_1) & k^l(\mathbf{x}_1, \mathbf{x}_2) & \dots & k^l(\mathbf{x}_1, \mathbf{x}_N) \\
        k^l(\mathbf{x}_2, \mathbf{x}_1) & k^l(\mathbf{x}_2, \mathbf{x}_2) & \dots & k^l(\mathbf{x}_2, \mathbf{x}_N) \\
        \vdots & \vdots & \ddots & \vdots \\
        k^l(\mathbf{x}_N, \mathbf{x}_1) & k^l(\mathbf{x}_N, \mathbf{x}_2) & \dots & k^l(\mathbf{x}_N, \mathbf{x}_N)
    \end{bmatrix} \otimes I_{D_l} \\ 
    & = \tilde{\Phi}^l_N \otimes I_{D_l},
\end{align*}
where $\otimes$ is the Kronecker product. The kernel function $k^l(\mathbf{x}_i, \mathbf{x}_j) \equiv \mathbb{E} \left[ \tilde{h}^l_{\boldsymbol{\cdot}}(\mathbf{x}_i) \tilde{h}^l_{\boldsymbol{\cdot}}(\mathbf{x}_j) \right]$ depends on the layer depth, the scale of the weights and biases and the amount of noise regularisation being applied. 
A schematic illustration of the covariance matrix is given in Figure \ref{fig: nggp covariance} for the simple case of only two inputs and two output units. To derive the elements of the covariance $\Phi^l$, consider the units $\tilde{h}^l_d, \tilde{h}^l_s$, $d, s \in \{1, ..., D_l\}$ and inputs $\mathbf{x}_i, \mathbf{x}_j, i,j \in \{1, ..., N\}$ which give 
\begin{align*}
    k^l_{ds}(\mathbf{x}_i, \mathbf{x}_j) & = \tilde{\Phi}^l_{ij} \otimes I_{ds} \\
    & = \begin{cases}
        \mathbb{E} \left[ \tilde{h}^{l}_d(\mathbf{x}_i) \tilde{h}^{l}_s(\mathbf{x}_j) \right]&, \text{ if } d=s \\
        0&, \text{otherwise}.
    \end{cases}
\end{align*}
Note that $k^l_{ds}(\mathbf{x}_i, \mathbf{x}_j) = 0, \forall i,j$ and $d \neq s$ due to the independence between the incoming connections (weights) associated with each output unit. Therefore, we only consider the case where $d=s$, which for $i \neq j$ gives 
\begin{align*}
     k^l_{d}&(\mathbf{x}_i,\mathbf{x}_j) = \mathbb{E}_{\mathbf{w}, b, \epsilon} \left[ \tilde{h}^{l}_d(\mathbf{x}_i) \tilde{h}^{l}_d(\mathbf{x}_j) \right] \\
    & = \frac{\sigma^2_w }{D_{l-1}} \sum^{D_{l-1}}_{d^{\prime}=1} \left[ \phi \left (\tilde{h}^{l-1}_{d^{\prime}}(\mathbf{x}_i) \right ) \phi \left ( \tilde{h}^{l-1}_{d^{\prime}}(\mathbf{x}_j) \right ) \right] + \sigma^2_b  \\
     & = \sigma^2_w\mathbb{E} \left[ \phi \left (\tilde{h}^{l-1}_{d^{\prime}}(\mathbf{x}_i) \right ) \phi \left ( \tilde{h}^{l-1}_{d^{\prime}}(\mathbf{x}_j) \right ) \right] + \sigma^2_b, 
\end{align*}
where the expectation is taking with respect to $\tilde{\mathbf{h}}^{l-1} \sim \mathcal{GP}(\mathbf{0}, \Phi^{l-1})$. 
The final equality follows from applying the above argument recursively for the previous layer $l-1$. 
For the case of $i = j$ (and $d=s$), we have that the diagonal components of the covariance matrix are given by
\begin{align*}
    k^l_{d}(\mathbf{x}_i, \mathbf{x}_i) = \sigma^2_w \left \{ \mathbb{E}\left [ \phi \left (\tilde{h}^{l-1}_{d^{\prime}}( \mathbf{x}_i) \right )^2\right ]\odot \mu_2 \right \}   + \sigma^2_b,
\end{align*}
where the influence of the noise $\epsilon$ is explicitly expressed through its second moment $\mu_2$. 
Using the initial condition
\begin{align*}
    k^0(\mathbf{x}_i, \mathbf{x}_j) = \mathbb{E}_\epsilon[(\mathbf{x}_i \odot \epsilon) \cdot (\mathbf{x}_j \odot \epsilon )]
\end{align*}
 and letting each layer width in the network $D_1, ..., D_L$ tend to infinity in succession, this recursive construction gives $\mathbf{f}$ as Gaussian distributed with mean and covariance
\begin{align}
    \label{eq: parameters of p of f}
    \mathbf{m} = \mathbf{0}, \textcolor{white}{space} K = \Phi^L. 
\end{align}
Finally, combining \eqref{eq: gp def}, \eqref{eq: conditional for y} and \eqref{eq: parameters of p of f} and using standard results for the marginal of a Gaussian distribution, the marginal in \eqref{eq: marginal of y} can be shown to be
\begin{align*}
    p(\mathbf{y}) = \mathcal{N}(\mathbf{0}, \Psi),
\end{align*}
where $\Psi(\mathbf{x}_i, \mathbf{x}_j) = K_{ij} + \sigma^2_\varepsilon \delta_{ij}$ with $\delta_{ij}$ the Kronecker delta \citep{williams2006gaussian}. Therefore, together with the additive noise level $\sigma^2_\varepsilon$, our model for the joint distribution over training outcomes is fully determined by the equivalent kernel corresponding to layer-wise recursion of an infinite basis function expansion. 
This kernel, in turn, depends on the parameterisation of the network and the amount of \textit{injected} noise. 


Having developed our noisy NNGP model, we next discuss predicting outcomes for unseen test data points. To make a prediction, we evaluate the predictive distribution $p(y^* | \mathbf{x}^*, \mathcal{D})$ at a new test point $\mathbf{x}^*$. Consider the joint
\begin{align*}
    p(\mathbf{y}, y^*| \mathbf{f}, \mathbf{x}^*) = \mathcal{N}(\mathbf{0}, \Psi^*)
\end{align*}
where we can partition the covariance $\Psi^*$ as follows
\begin{align*}
    \Psi^* = \begin{bmatrix}
        \Psi & \mathbf{k} \\
        \mathbf{k}^T & \psi 
    \end{bmatrix}
\end{align*}
with $\mathbf{k} = [k^L_1(\mathbf{x}_1, \mathbf{x}^*), ..., k^L_1(\mathbf{x}_N, \mathbf{x}^*), ..., k^L_{D_L}(\mathbf{x}_N, \mathbf{x}^*)]^T$, an $ND_L$ dimensional vector and $\psi = k^L(\mathbf{x}^*, \mathbf{x}^*) + \sigma^2_\varepsilon$. 
Using standard results for the conditional distribution of a partitioned Gaussian vector, we find
\begin{align}
    \label{eq: predictive distribution}
    p(y^* | \mathbf{f}, \mathbf{x}^*, \mathbf{y}) = \mathcal{N}(\mu(\mathbf{x}^*), \sigma^2(\mathbf{x}^*))
\end{align}
where $\mu(\mathbf{x}^*) = \mathbf{k}^T\Psi^{-1}\mathbf{y}$ and $\sigma^2(\mathbf{x}^*) = \psi - \mathbf{k}^T\Psi^{-1}\mathbf{k}$. 
This result is the function space equivalent to exact Bayesian inference in parameter space: 
by computing the conditional in \eqref{eq: predictive distribution}, we are implicitly performing an integration over the posterior of the parameters associated with an infinitely wide noise regularised neural network \citep{williams1997computing}. 

In the next section, we study how the properties of the kernel derived in this section depend the parameters of the network when using ReLU activations. 
Furthermore, for the remainder of the paper we drop the dependence on the hidden units and training set indices for ease of notation and simply refer to $k^l_{d}(\mathbf{x}_i, \mathbf{x}_j)$ as $k^l(\mathbf{x}, \mathbf{x}^{\prime})$.  


\section{Kernel parameters and critical neural network initialisation}

\begin{table*}[t]
	\caption{Limiting behaviour for degenerate and critical noisy ReLU kernels.}
	\label{tab: limiting kernel}
	\begin{center}
		\begin{small}
			\begin{sc}
				\begin{adjustbox}{width=0.8\linewidth,center}
				\begin{tabular}{lll}
					\toprule
					Weight variance & Bias variance &  Limiting value as $L \rightarrow \infty$                                                                                       \\
					\toprule
					- Additive Noise ($\mu_2 > 0$) -  \\
					\midrule
					a.1) $0 \leq \sigma^2_w < 2$ & $\sigma^2_b \geq 0$ & $k^L(\mathbf{x}, \mathbf{x}) \rightarrow a \text{ (const. independent of $\mathbf{x}$) }$ \\
					a.2) $\sigma^2_w \geq 2$ & $\sigma^2_b \geq 0$ & $k^L(\mathbf{x}, \mathbf{x}) \rightarrow \infty$ \\
					\toprule
					- Mult. Noise ($\mu_2 > 1$) -  \\
					\midrule
					m.1) $0 \leq \sigma^2_w < 2/\mu_2$ & $\sigma^2_b = 0$ & $k^L(\mathbf{x}, \mathbf{x}) \rightarrow 0$ \\
					m.2) $0 \leq \sigma^2_w < 2/\mu_2$ & $\sigma^2_b > 0$ & $k^L(\mathbf{x}, \mathbf{x}) \rightarrow a \text{ (const. independent of $\mathbf{x}$) }$ \\ 
    				m.3) $\sigma^2_w > 2/\mu_2$ & $\sigma^2_b \geq 0$ & $k^L(\mathbf{x}, \mathbf{x}) \rightarrow \infty$ \\
					m.4) $\sigma^2_w = 2/\mu_2$ & $\sigma^2_b > 0$ & $k^L(\mathbf{x}, \mathbf{x}) \rightarrow \infty$ \\
					m.5) $\sigma^2_w = 2/\mu_2$ & $\sigma^2_b = 0$ & $k^L(\mathbf{x}, \mathbf{x}) \rightarrow k^{L-1}(\mathbf{x}, \mathbf{x}) = ... = k^0(\mathbf{x}, \mathbf{x})$ \\
					\bottomrule
				\end{tabular}
			\end{adjustbox}
			\end{sc}
		\end{small}
	\end{center}
	\vskip -0.1in
\end{table*}

We begin by examining the interaction between the parameters of the noisy NNGP kernel and its corresponding network initialisation. Specifically, we focus on ReLU activations and show that the kernel parameter values that lead to non-degenerate kernels for deep noisy NNGPs are exactly those that promote stable signal propagation in noise regularised ReLU networks. 

Let $\phi(a) = \textrm{ReLU}(a) = \max(0, a)$ and define
\begin{align*}
    \rho^l_{\mathbf{x}, \mathbf{x}^{\prime}} = \frac{k^l(\mathbf{x}, \mathbf{x}^{\prime})}{\sqrt{k^l(\mathbf{x}, \mathbf{x})k^l(\mathbf{x}^{\prime}, \mathbf{x}^{\prime})}},
\end{align*}
then the elements of the covariance $\Phi^l$ at a hidden unit are
\begin{align}
	\label{eq: kernel covariance}
    k^l(\mathbf{x}, \mathbf{x}^{\prime}) = \frac{\sigma^2_w}{2}k^{l-1}(\mathbf{x}, \mathbf{x}^{\prime})\left \{ g(\rho^{l-1}_{\mathbf{x}, \mathbf{x}^{\prime}}) + \frac{1}{2} \right \} + \sigma^2_b,
\end{align}
for $\mathbf{x} \neq \mathbf{x}^\prime$, where 
\begin{align*}
	g(\rho^{l}_{\mathbf{x}, \mathbf{x}^{\prime}}) = \frac{\rho^{l}_{\mathbf{x}, \mathbf{x}^{\prime}}\sin^{-1} \left (\rho^{l}_{\mathbf{x}, \mathbf{x}^{\prime}} \right ) + \sqrt{1 - (\rho^{l}_{\mathbf{x}, \mathbf{x}^{\prime}})^2}}{\pi\rho^{l}_{\mathbf{x}, \mathbf{x}^{\prime}}},
\end{align*}
with diagonal elements given by
\begin{align}
    \label{eq: kernel variance}
    k^l(\mathbf{x}, \mathbf{x}) = \frac{\sigma^2_w}{2}  k^{l-1}(\mathbf{x}, \mathbf{x}) \odot \mu_2    + \sigma^2_b.
\end{align}
These formulae are the kernel equivalent to the signal propagation recurrences derived in \citep{pretorius2018critical} for noisy ReLU networks. For no noise and outside the context of GPs, a similar result can be found in \citep{cho2009kernel}. Repeated substitution in \eqref{eq: kernel variance} shows that
\begin{align}
	\label{eq: expanded kernel variance}
    & k^L(\mathbf{x}, \mathbf{x}) = \frac{\sigma^2_w}{2}  \left ( \frac{\sigma^2_w}{2}  k^{L-2}(\mathbf{x}, \mathbf{x}) \odot \mu_2    + \sigma^2_b  \right) \odot \mu_2  + \sigma^2_b \nonumber \\
    & \textcolor{white}{d} \vdots \nonumber \\
    & = \begin{cases}
        \left(\frac{\sigma^2_w}{2}\right)^L k^{0}(\mathbf{x}, \mathbf{x}) + \sum^{L-1}_{l=0}\left(\frac{\sigma^2_w}{2}\right)^l (\mu_2 + \sigma^2_b), \\ \hspace{0.8cm} \textcolor{white}{s} \text{if } \odot \equiv + \text{ (Additive noise)}, \\
        \left(\frac{\sigma^2_w\mu_2}{2}\right)^L k^{0}(\mathbf{x}, \mathbf{x}) + \sum^{L-1}_{l=0}\left(\frac{\sigma^2_w\mu_2}{2}\right)^l \sigma^2_b, \\ \hspace{1cm} \text{if } \odot \equiv \times \text{ (Multiplicative noise)}.
    \end{cases}
\end{align}

The limiting properties of the kernel are seen by letting $L \rightarrow \infty$ in \eqref{eq: expanded kernel variance}. 
In this limit, several degenerate kernels arise, analogous to cases of unstable signal propagation dynamics in mean-field theory and other related work \citep{poole2016exponential, daniely2016toward, schoenholz2016deep, pretorius2018critical}. 
We provide the different cases in Table \ref{tab: limiting kernel}. 
For any amount of additive noise, all possible settings (see A.1 and A.2) of the kernel parameters $\sigma^2_w$ and $\sigma^2_b$ in \eqref{eq: expanded kernel variance} will result in a degenerate kernel in the limit of infinite depth. 
The situation is similar for multiplicative noise, except for the case (M.5), where $\{\sigma^2_w, \sigma^2_b\} = \{2/\mu_2, 0\}$. We refer to these parameters in (M.5) as the \textit{critical kernel parameters}. 
Here, the diagonal elements of the covariance stay fixed at their initial values even at extreme depth. 
These parameter values are identical to the proposed initialisations for deep noisy ReLU networks derived in \citep{pretorius2018critical}.

From \eqref{eq: kernel covariance} we can see that the off-diagonal elements of the covariance matrix are influenced by the noise level at the critical values through the relationship $\sigma^2_w/2 = 1/\mu_2$. 
Furthermore, we note that $k^l(\mathbf{x}, \mathbf{x}^{\prime}) \rightarrow 0$ as $\mu_2 \rightarrow \infty$.    
Therefore, multiplicative noise regularisation has a damping effect on the kernel function evaluated between different inputs, tending towards total dissimilarity and a diagonal covariance. This reduction in the richness of the covariance structure exploitable by the NNGP then enforces a strong prior for simple functions away from the training points. 
To see this effect, consider the predictive distribution in \eqref{eq: predictive distribution}, for a test point $\mathbf{x}^*$. 
For large amounts of noise, $\mathbf{k} \rightarrow \mathbf{0}$ and therefore in the limit, $\mu(\mathbf{x}^*) = 0$ and $\sigma^2(\mathbf{x}^*) = \psi$. 
Since $\Psi$ is symmetric positive definite by definition and $\mathbf{k}^T\Psi^{-1}\mathbf{k} \geq 0, \forall \mathbf{k}$, the predicted outcome $y^*$ will be a sample from a zero mean Gaussian with maximal uncertainty as measured by the variance $\psi$, i.e $y^*|\mathbf{x}^* \sim \mathcal{N}(0, \psi)$. 

To validate the above claims, the following section provides an empirical investigation. In particular, we test two hypotheses that stem from the above theoretical analysis, using both real-world and synthetic datasets. 





\section{Experiments}

We have shown how the kernel parameters for a noisy NNGP relate to those for its corresponding deep neural network. 
In doing so, our discussion has led us to the following testable hypotheses:

\begin{figure*}
    \centering
    \includegraphics[width=0.7\linewidth]{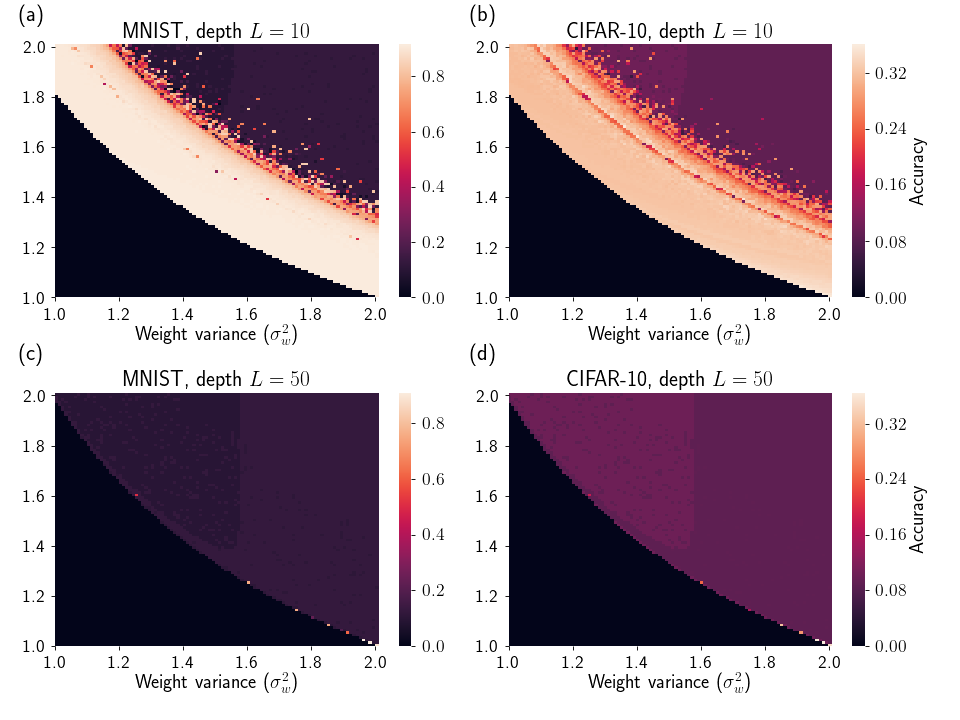}
	\caption{\textit{Sensitivity of NNGP kernel parameters for different depths.} \textbf{(a)} Test accuracy for NNGP with depth $L = 10$ on MNIST with training and test set sizes of $N=1000$ and kernel parameters $\sigma^2_b = 0$, $\sigma^2_w = 1.0$ to $2.0$ and $\mu_2 = 1.0$ to $2.0$, both equally spaced using interval sizes of $0.01$. \textbf{(b)} NNGP with depth $L=10$ on CIFAR-10 ($N=1000$). \textbf{(c) - (d)} Same as (a) and (b) but with depth $L = 50$.  } 
	\label{fig: nggp performance 2}
\end{figure*}

\begin{figure*}
    \centering
    \includegraphics[width=\linewidth]{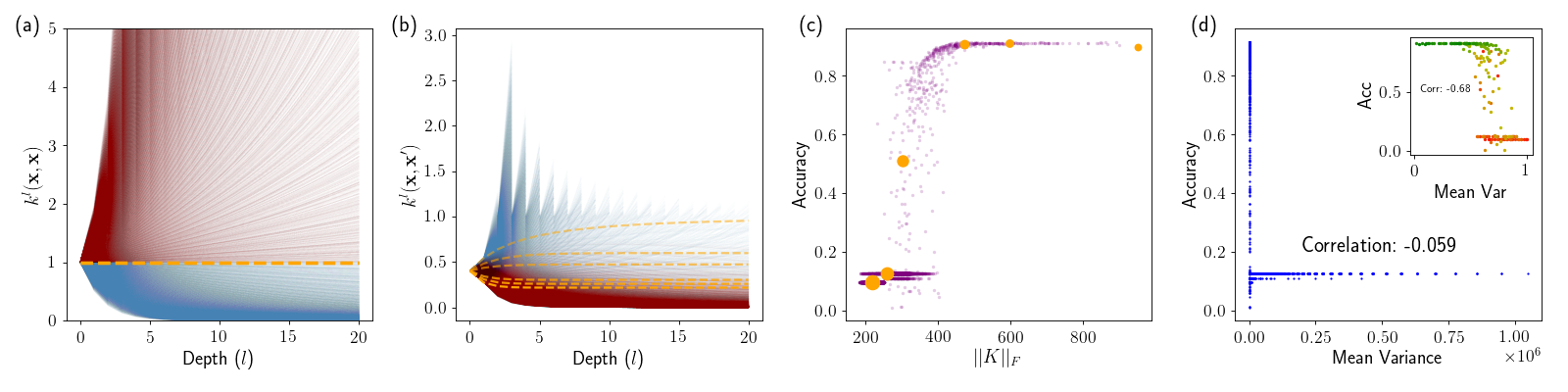}
    \vspace{-0.2cm}
	\caption{\textit{Effects of noise induced regularisation on the noisy NNGPs in Figure \ref{fig: nggp performance} for MNIST.} \textbf{(a)} Depth evolution of $k^l(\mathbf{x}, \mathbf{x})$ for the different kernel parameters $\sigma^2_w = 2/\mu_2$ (dashed orange lines), $\sigma^2_w > 2/\mu_2$ (solid red lines) and $\sigma^2_w < 2/\mu_2$ (solid blue lines), with $\sigma^2_b = 0$ in all experiments.  \textbf{(b)} Depth evolution of $k^l(\mathbf{x}, \mathbf{x}^{\prime})$ \textbf{(c)} Relationship between accuracy and covariance norm. Orange points correspond to critical kernel parameters with larger sizes indicating more noise. \textbf{(d)} Scatter plot of accuracy as a function of mean variance. We measure the quality of uncertainty estimates by computing the correlation of the mean posterior predictive variance with test accuracy. Main plot contains all points, whereas the inset only contains points close to criticality (green to red showing an increase in noise).} 
	\label{fig: noise induced reg in nngps}
\end{figure*}

\begin{figure*}
    \centering
    \includegraphics[width=0.8\linewidth]{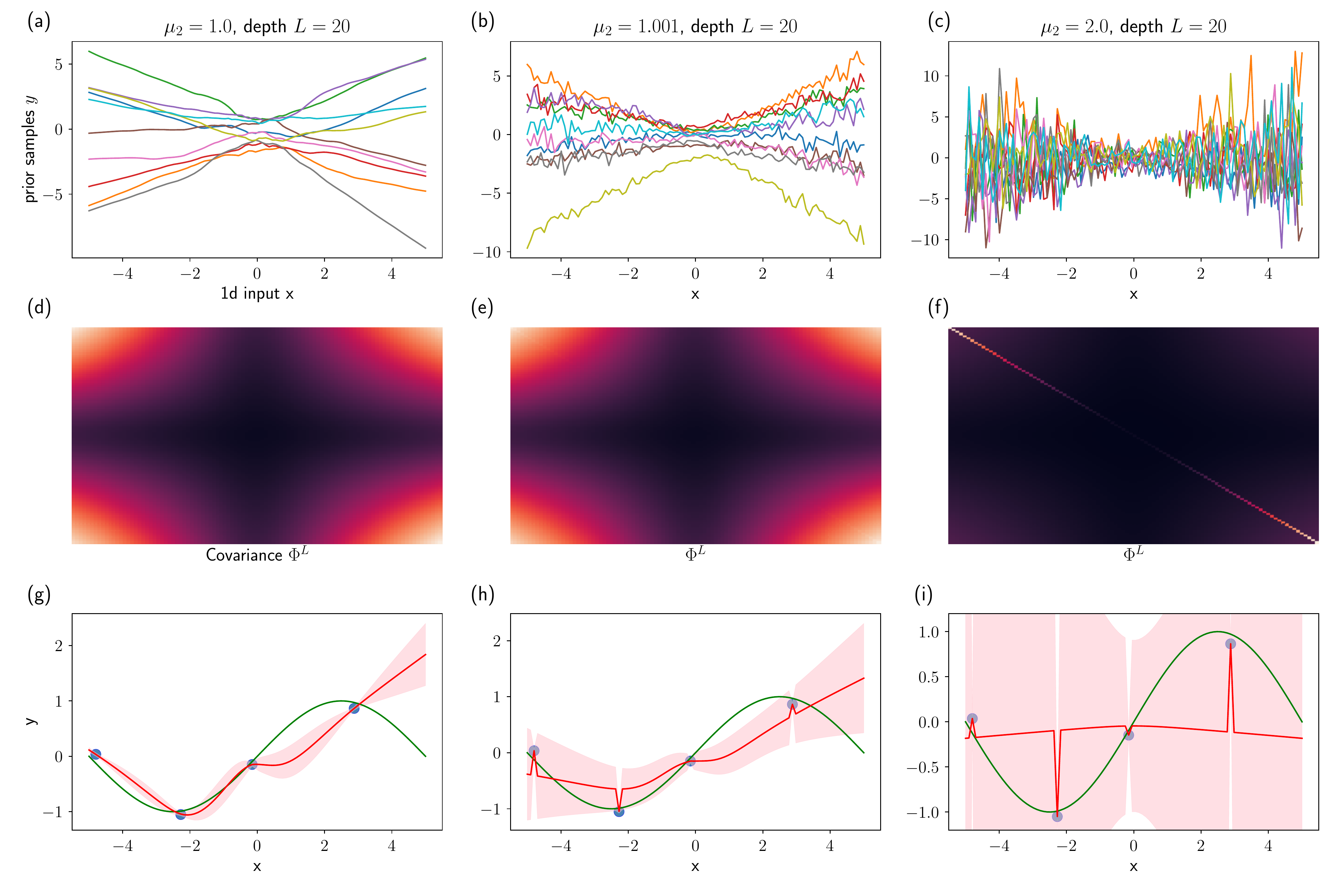}
    \vspace{-0.5cm}
	\caption{\textit{20-layer noisy NNGPs with 1D input data and $\{\sigma^2_w, \sigma^2_b\} = \{2/\mu_2, 0.05\}$}: Left column $\mu_2 = 1.0$, middle column $\mu_2 = 1.001$ and right column $\mu_2 = 2.0$. \textbf{(a) - (c)} Samples from NNGP prior. \textbf{(d) - (f)} NNGP covariance $\Phi^L$, $L=20$. \textbf{(g) - (i)} NNGP fit (red line) using four training examples (blue dots) sampled from a simple sinusoidal function (green line) with $\sigma_\varepsilon = 0.1$. } 
	\label{fig: 1d nngps}
\end{figure*}

\begin{enumerate}
    \item[\textbf{H1-}] \textbf{Noisy NNGPs perform best at their critical parameters}: The sensitivity of the kernel parameters  should cause noisy NNGPs to perform poorly at settings far away from the critical kernel parameters. 
    Furthermore, the reliance on these critical values for good performance should become more marked at greater depth [Figures \ref{fig: nggp performance} and \ref{fig: nggp performance 2}].
    \item[\textbf{H2-}] \textbf{Noise constrains the covariance and leads to simpler models away from the training points with larger uncertainty}: Even at criticality, noise injection applies a shrinkage effect to the kernel function evaluated between different inputs to the noisy NNGP. 
    This should lead to a constrained covariance structure, or in the limit of a large amount of noise, a completely diagonal covariance. 
    The NNGP prior over functions regularised in this way should lead to simpler models away from the training points with increased estimates of uncertainty [Figures \ref{fig: noise induced reg in nngps} and \ref{fig: 1d nngps}].
\end{enumerate}

\textbf{H1}: We begin by investigating the sensitivity of the kernel parameter values. 
As shown in Figure \ref{fig: nggp performance}, we test the performance of $20$-layer NNGPs on MNIST and CIFAR-10 with kernels constructed for a grid of variance parameters $\sigma^2_w = 1.0, 1.01, ..., 1.99, 2.0$, for varying values of the noise level parameter $\mu_2 = 1.0, 1.01, ..., 1.99, 2.0$. 
Our approach to classification in this paper is identical to \citep{lee2017deep}, where classification is treated as a regression problem.
Specifically, instead of one-hot output vectors, each output vector is recoded as a zero mean regression output with the value $0.9$ in the index corresponding to the correct class and $-0.1$ for all other indices corresponding to the incorrect classes. 
The predicted class label for a given input is then simply the index corresponding to the maximum value in the output vector as predicted by the NNGP regression model.
For all experiments, we set $\sigma^2_b = 0$. 
Figures \ref{fig: nggp performance}(b) and (c) confirm our expectations, showing that the kernel parameters corresponding to NNGPs with good performance closely follow the critical initialisation boundary $\sigma^2_w = 2/\mu_2$ shown in Figure \ref{fig: nggp performance}(a). 
As kernels are constructed further away from criticality, their performances start to deteriorate. 

The sensitivity to the kernel parameters becomes more acute at larger depth as shown in Figure \ref{fig: nggp performance 2}. 
Panels (a) and (b) plot the results for a shallower depth of $L = 10$. 
In this case, a wide band is seen to form around the critical boundary (beige shaded area) with kernel parameters far away from their critical values still able to perform reasonably well. 
This is no longer the case in Panels (c) and (d), where we tested performances at a greater depth, $L=50$. 
At this depth, the NNGP is far more sensitive to the kernel parameters and only a few models with kernel parameters very close to the critical boundary are seen to perform well.  

\textbf{H2}: For all the models evaluated in H1, we also study the influence of the noise on the kernel as well as on the resulting NNGP covariance matrix. 
For each model, we plot in Figure \ref{fig: noise induced reg in nngps}(a) and (b) the depth evolution of the kernel, using two inputs from the MNIST dataset. In (a) we track the variance of one of the inputs and in (b) the covariance between the two inputs. 
The dashed orange lines correspond to the kernel parameters $\sigma^2_w = 2/\mu_2$, with $\sigma^2_w > 2/\mu_2$ shown in solid red and $\sigma^2_w < 2/\mu_2$ shown in solid blue. (Recall that $\sigma^2_b = 0$ for all experiments). 
The limiting behaviour described in (M.1), (M.3) and (M.5) in Table \ref{tab: limiting kernel} is shown in (a), with all kernels tending towards degenerate function mappings, except those evolving under the critical parameters. 
Furthermore, in (b), we show the damping effect on the kernel at criticality, highlighted by decreasing asymptotes (dashed orange lines around layer 20) as more noise is being applied (darker lines).

The depth dynamics of the kernel also constrains the resulting covariance matrix. 
To see the effect of this, we use the Frobenius norm of the covariance matrix as a proxy for its richness. 
Figure \ref{fig: noise induced reg in nngps}(c) plots the relationship between the covariance norm and test accuracy for all the experiments in H1. Interestingly, the norm seems to suggest a step function relationship. 
Moving from right to left in (c), we observe a sudden and dramatic drop in performance beyond a certain amount of regularisation, as measured by a decreasing covariance norm. 
In other words, there seems to exist some requisite amount of information to be captured by the covariance matrix in order for the NNGP to be able to perform well. 
This is also the case at criticality: in (c), the orange points correspond to critical kernels with larger points associated with more noise. 

The effect of increased noise regularisation on uncertainty estimation is shown in Figure \ref{fig: noise induced reg in nngps}(d), where we plot test accuracy as a function of the mean posterior predictive variance. 
For NNGPs far away from criticality (blue points in main plot), we see little correlation ($-0.059$) between variance and test accuracy. 
The inset in (d) shows the same plot but for NNGPs close to their critical parameters. 
For these models the (negative) correlation is stronger ($-0.68$), possibly providing more reliable uncertainty estimates. 
Here, the green points are low noise models and the red points are high noise models. 
As expected, noise regularisation causes the posterior predictive variance to increase leading to higher uncertainty. 
We did the same investigations using the CIFAR-10 dataset and obtained similar results (see Appendix A).

Finally, to gain more insight, we consider a simple one-dimensional regression task.\footnote{The example is taken from Chapter 1 in \cite{bishop2006pattern}.} 
The top row in Figure \ref{fig: 1d nngps} shows samples from a 20-layer NNGP prior for (a) $\mu_2 = 1.0$ (no noise), (b) $\mu_2 = 1.001$ (small noise), (c) $\mu_2 = 2.0$ (large noise) and $\{\sigma^2_w, \sigma^2_b\} = \{2/\mu_2, 0.05\}$.  We found a small amount of bias $\sigma^2_b =0.05$, improved each fit (see Appendix B for a discussion on non-zero biases). The covariance structure corresponding to each NNGP is shown in (d)-(f), located in the middle row of Figure \ref{fig: 1d nngps}. 
The bottom row, (g)-(i) shows the fit from the posterior predictive (red line) using four training examples (blue dots) sampled from a simple sinusoidal function (green line) with $\sigma_\varepsilon = 0.1$. 
Moving across the columns from left to right, we find that the samples from the prior become more erratic as the covariance becomes diagonal, which strongly regularises the regression model at previously unseen test points. 
Note that the model in (i) still corresponds to exact Bayesian inference, but with a strong prior for near constant functions with high uncertainty away from the training points.


\section{Discussion}

We have shown that critical initialisation of noisy ReLU networks corresponds to a choice of optimal kernel parameters in noisy NNGPs and that deviation from these critical parameters leads to poor performance, becoming more severe with depth and the extent of the deviation.
In addition, we highlighted the effect of noise on the covariance of a noisy NNGP at criticality, with noise in the limit yielding a fully diagonal covariance, acting as a regulariser on the posterior predictive.



It is interesting to reflect on the connection between deep NNs and GPs in the context of representation learning and noise regularisation.
The core assumption in deep learning is that deep NNs learn distributed hierarchical representations useful for modeling the true underlying data generating mechanism, whereas shallow models do not. 
In these deeper models, noise regularisation is thought to be successful largely because of its influence on representations at different levels of abstraction during the training procedure \citep{goodfellow2016deep}.
As discussed in previous work \citep{neal1994priors, matthews2018gaussian}, the kernels associated with NNGPs do not use learned hierarchical representations.
Nevertheless these models are still able to perform as well, or sometimes better than, their neural network counterparts \citep{lee2017deep}.
In the infinite width setting, the success of regularisation from noise injection is unlikely to have the same interpretation as in the finite width setting.
We note that in the context of NNGPs, noise regularisation has a stronger connection with controlling the length scale parameter in commonly used kernel functions than regularising through corrupted representations at different levels of abstraction.  
This connection with the length scale parameter means that noise regularisation in NNGPs may be more accurately interpreted as a useful mechanism to designing priors by controlling the smoothness of the kernel function.

Finally, recent work related to NNGPs, has made it possible to accurately model the learning dynamics of deep neural networks by taking a function space perspective of gradient descent training in the infinite width limit \citep{jacot2018neural, lee2019wide}. 
We envision a similar analysis could be applied to accurately model the learning dynamics of noise regularised deep neural networks.

\clearpage
\newpage

\bibliographystyle{IEEEtranN}
\bibliography{bibfile}

\clearpage
\newpage

\appendix
\section*{Appendix}
\addcontentsline{toc}{section}{Appendices}
\renewcommand{\thesubsection}{\Alph{subsection}}

\subsection{Additional results}

\begin{figure*}
    \centering
    \includegraphics[width=0.9\linewidth]{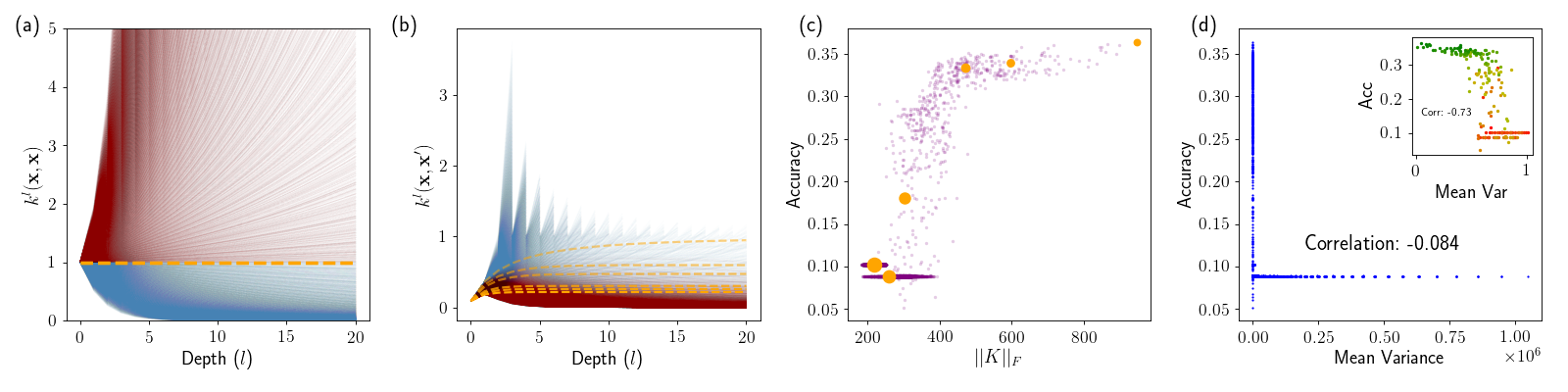}
    \vspace{-0.2cm}
	\caption{\textit{Effects of noise induced regularisation on NNGPs in Figure 1 for CIFAR-10.} \textbf{(a)} Depth evolution of $k^l(\mathbf{x}, \mathbf{x})$ for the different kernel parameters $\sigma^2_w = 2/\mu_2$ (dashed orange lines), $\sigma^2_w > 2/\mu_2$ (solid red lines) and $\sigma^2_w < 2/\mu_2$ (solid blue lines), with $\sigma^2_b = 0$ in all experiments.  \textbf{(b)} Depth evolution of $k^l(\mathbf{x}, \mathbf{x}^{\prime})$ \textbf{(c)} Relationship between accuracy and covariance norm. Orange points correspond to critical kernel parameters with larger sizes indicating more noise. \textbf{(d)} Quality of uncertainty estimates as measured by the correlation of the mean posterior predictive variance with test accuracy. Main plot contains all points, whereas the inset only contains points close to criticality (green to red showing an increase in noise).} 
	\label{fig: noise induced reg in nngps cifar 10}
\end{figure*}

Figure \ref{fig: noise induced reg in nngps cifar 10} provides additional results using CIFAR-10 instead of MNIST for the experiments presented in Figure 1. The results are similar to those described in the main text for MNIST.

\subsection{Kernels with non-zero biases}

In our experiments, we noticed that noisy ReLU NNGPs often benefit from small non-zeros biases. Therefore, we consider here the implication of non-zero biases for the evolution of the diagonal terms in the NNGP covariance. Recall that the diagonal (variance) terms of the covariance matrix can be expanded as follows
\begin{align}
	\label{eq: expanded kernel variance suppmat}
	& k^L(\mathbf{x}, \mathbf{x}) = \frac{\sigma^2_w}{2}  \left ( \frac{\sigma^2_w}{2}  k^{L-2}(\mathbf{x}, \mathbf{x}) \odot \mu_2    + \sigma^2_b  \right) \odot \mu_2  + \sigma^2_b \nonumber \\
    & \textcolor{white}{d} \vdots \nonumber \\
    & = \begin{cases}
        \left(\frac{\sigma^2_w}{2}\right)^L k^{0}(\mathbf{x}, \mathbf{x}) + \sum^{L-1}_{l=0}\left(\frac{\sigma^2_w}{2}\right)^l (\mu_2 + \sigma^2_b), \\ \hspace{0.8cm} \textcolor{white}{s} \text{if } \odot \equiv + \text{ (Additive noise)}, \\
        \left(\frac{\sigma^2_w\mu_2}{2}\right)^L k^{0}(\mathbf{x}, \mathbf{x}) + \sum^{L-1}_{l=0}\left(\frac{\sigma^2_w\mu_2}{2}\right)^l \sigma^2_b, \\ \hspace{1cm} \text{if } \odot \equiv \times \text{ (Multiplicative noise)}.
    \end{cases}
\end{align}
We first focus on the multiplicative noise case, at the critical weight variance $\sigma^2_w = \frac{2}{\mu_2}$. Here, a non-zero bias translates into a second term $(L-1)\sigma^2_b$ in \eqref{eq: expanded kernel variance suppmat}, that grows linearly with depth. For small initialised biases in typically deep networks this term will be small. For example, a $20$-layer deep neural network with $\sigma^2_b = 0.05$, will translate into an NNGP covariance matrix with diagonal covariance terms given by $k^{0}(\mathbf{x}, \mathbf{x}) + 1$. Therefore, at depth, the linear growth in a non-zero $\sigma^2_b$, is far less severe than the exponential growth/decay from an incorrect setting of $\sigma^2_w$. In the additive noise case with $\sigma^2_w = 2$, the situation is similar, but with an added linear growth in noise. Unfortunately, it is less straightforward to analyse the effects of non-zero biases on the off-diagonal covariance terms. 



\end{document}